\documentclass[11pt,a4paper]{article}
\usepackage[hyperref]{emnlp-ijcnlp-2019}
\usepackage{times}
\usepackage{latexsym}
\usepackage{amsmath}
\usepackage{enumitem}
\usepackage{url}
\usepackage{graphicx}
\usepackage{amsfonts}
\usepackage{booktabs}
\usepackage{siunitx}
\usepackage{setspace}
\usepackage{float} 
\usepackage{xcolor}
\aclfinalcopy 

\DeclareMathOperator*{\argmax}{arg\,max}

\graphicspath{ {./images/} }

\definecolor{light-gray}{gray}{0.95}
\newcommand{\code}[1]{{\texttt{#1}}}

\title{Let Me Know What to Ask: Interrogative-Word-Aware Question Generation}
\author{Junmo Kang\thanks{\ \ Equal contribution.}\qquad Haritz Puerto San Roman\footnotemark[1]  \qquad Sung-Hyon Myaeng\\
School of Computing, KAIST \\ 
Daejeon, Republic of Korea\\ 
\{junmo.kang, haritzpuerto94, myaeng\}@kaist.ac.kr}

\date{}

\begin{document}
\maketitle
\begin{abstract}
Question Generation (QG) is a Natural Language Processing (NLP) task that aids advances in Question Answering (QA) and conversational assistants. Existing models focus on generating a question based on a text and possibly the answer to the generated question. They need to determine the type of interrogative word to be generated while having to pay attention to the grammar and vocabulary of the question. In this work, we propose Interrogative-Word-Aware Question Generation (IWAQG), a pipelined system composed of two modules: an interrogative word classifier and a QG model. The first module predicts the interrogative word that is provided to the second module to create the question. Owing to an increased recall of deciding the interrogative words to be used for the generated questions, the proposed model achieves new state-of-the-art results on the task of QG in SQuAD, improving from 46.58 to 47.69 in BLEU-1, 17.55 to 18.53 in BLEU-4, 21.24 to 22.33 in METEOR, and from 44.53 to 46.94 in ROUGE-L.
\end{abstract}

\section{Introduction}
Question Generation (QG) is the task of creating questions about a text in natural language. This is an important task for Question Answering (QA) since it can help create QA datasets. It is also useful for conversational systems like Amazon Alexa. Due to the surge of interests in these systems, QG is also drawing the attention of the research community. One of the reasons for the fast advances in QA capabilities is the creation of large datasets like SQuAD \cite{Rajpurkar_2016} and TriviaQA \cite{Joshi_2017}. Since the creation of such datasets is either costly if done manually or prone to error if done automatically, reliable and meaningful QG can play a key role in the advances of QA \cite{lewis2019unsupervisedqa}.

\begin{figure}
    \centering
    \includegraphics[width=7.5cm]{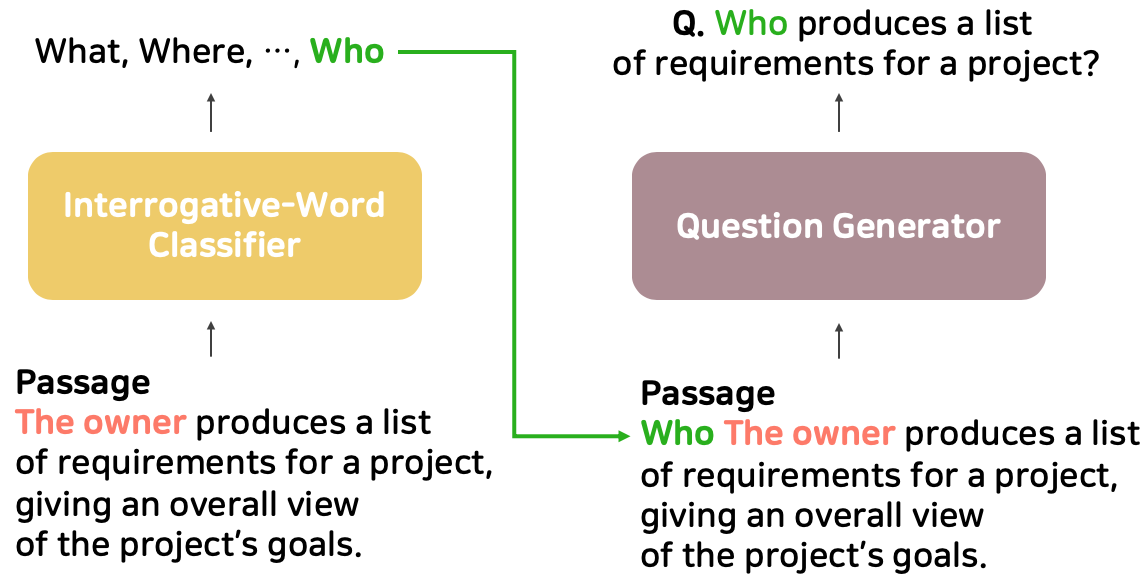}
    \caption{High-level overview of the proposed model.}
    \label{fig:architecture}
\end{figure}{}

QG is a difficult task due to the need of understanding the text to ask about and generating a natural question that is adequate according to the given text. We consider that this task have two aspects: \textit{what to ask} and \textit{how to ask}. The first one refers to the information about the entity that we want to ask; this includes the interrogative word to use and the topic of the question. On the other hand, \textit{how to ask} refers to the creation of a natural language question that is grammatically correct and semantically precise. Most of the current approaches utilize sequence-to-sequence models, composed of an encoder model that first transforms a passage into a vector and a decoder model that given this vector, generates a question about the passage \cite{Liu:2019:LGQ:3308558.3313737, sun-etal-2018-answer, zhao2018paragraph, pan2019recent}. 

There are different settings for QG. \citet{subramanian2018neural} assumes that only a passage is given and attempts to find candidate key phrases that represent the core of the questions to be created. \citet{zhao2018paragraph} follows an answer-aware setting, where the input is a passage and the answer to the question to create. We assume this setting and consider that the answer is a span of the passage, as in SQuAD. Following this approach, the decoder of the sequence-to-sequence model has to learn to generate both the interrogative word (i.e., wh-word) and the rest of the question simultaneously.

The main claim of our work is that separating the two tasks (i.e., interrogative-word classification and question generation) can lead to a better performance. We posit that the interrogative word must be predicted by a well-trained classifier. We consider that selecting the right interrogative word is the key to generate high-quality questions. For example, a question with a wrong interrogative word for the answer ``the owner" is: ``what produces a list of requirements for a project?". However, with the right interrogative word, \textit{who}, the question would be: ``who produces a list of requirements for a project?", which is clear that is more adequate regarding the answer than the first one. According to our claim, the independent classification model can improve the recall of interrogative words of a QG model because 1) the interrogative word classification task is easier to solve than generating the interrogative word along with the full question in the QG model and 2) the QG model would be able to generate the interrogative word easily by using the copy mechanism, which can copy parts of the input of the encoder. With these hypotheses, we propose Interrogative-Word-Aware Question Generation (IWAQG), a pipelined system composed of two modules: an interrogative-word classifier that predicts the interrogative word and a QG model that generates a question conditioned on the predicted interrogative word. Figure \ref{fig:architecture} shows a high-level overview of our approach.

The proposed model achieves new state-of-the-art results on the task of QG in SQuAD, improving from 46.58 to 47.69 in BLEU-1, 17.55 to 18.53 in BLEU-4, 21.24 to 22.33 in METEOR, and from 44.53 to 46.94 in ROUGE-L.

\section{Related Work}
Question Generation (QG) problem has been approached in two ways. One is based on heuristics, templates and syntactic rules \cite{heilman-smith-2010-good, qgrules, Labutov2015DeepQW}. This type of approach requires a heavy human effort, so they do not scale well. The other approach is based on neural networks and it is becoming popular due to the recent progress of deep learning in NLP \cite{pan2019recent}. \citet{du2017learning} is the first one to propose an sequence-to-sequence model to tackle the QG problem and outperformed the previous state-of-the-art model using human and automatic evaluations.

\citet{sun-etal-2018-answer} proposed a similar approach to us, an answer-aware sequence-to-sequence model with a special decoding mode in charge of only the interrogative word. However, we propose to predict the interrogative word before the encoding stage, so that the decoder can focus more on the rest of the question rather than on the interrogative word. Besides, they cannot train the interrogative-word classifier using golden labels because it is learned implicitly inside the decoder. \citet{Duan2017QuestionGF} proposed, in a similar way to us, a pipeline approach. First, the authors create a long list of question templates like ``who is author of", and ``who is wife of". Then, when generating the question, they select first the question template and next, they fill it in. To select the question template, they proposed two approaches. One is a retrieval-based question pattern prediction, and the second one is a generation-based question pattern prediction. The first one has the problem that is computationally expensive when the question pattern size is large, and the second one, although it yields to better results, it is a generative approach and we argue that just modeling the interrogative word prediction as a classification task is easier and can lead to better results. As far as we know, we are the first one to propose an explicit interrogative-word classifier that provides the interrogative word to the question generator.

\begin{figure*}
    \centering
    \includegraphics[width=\textwidth]{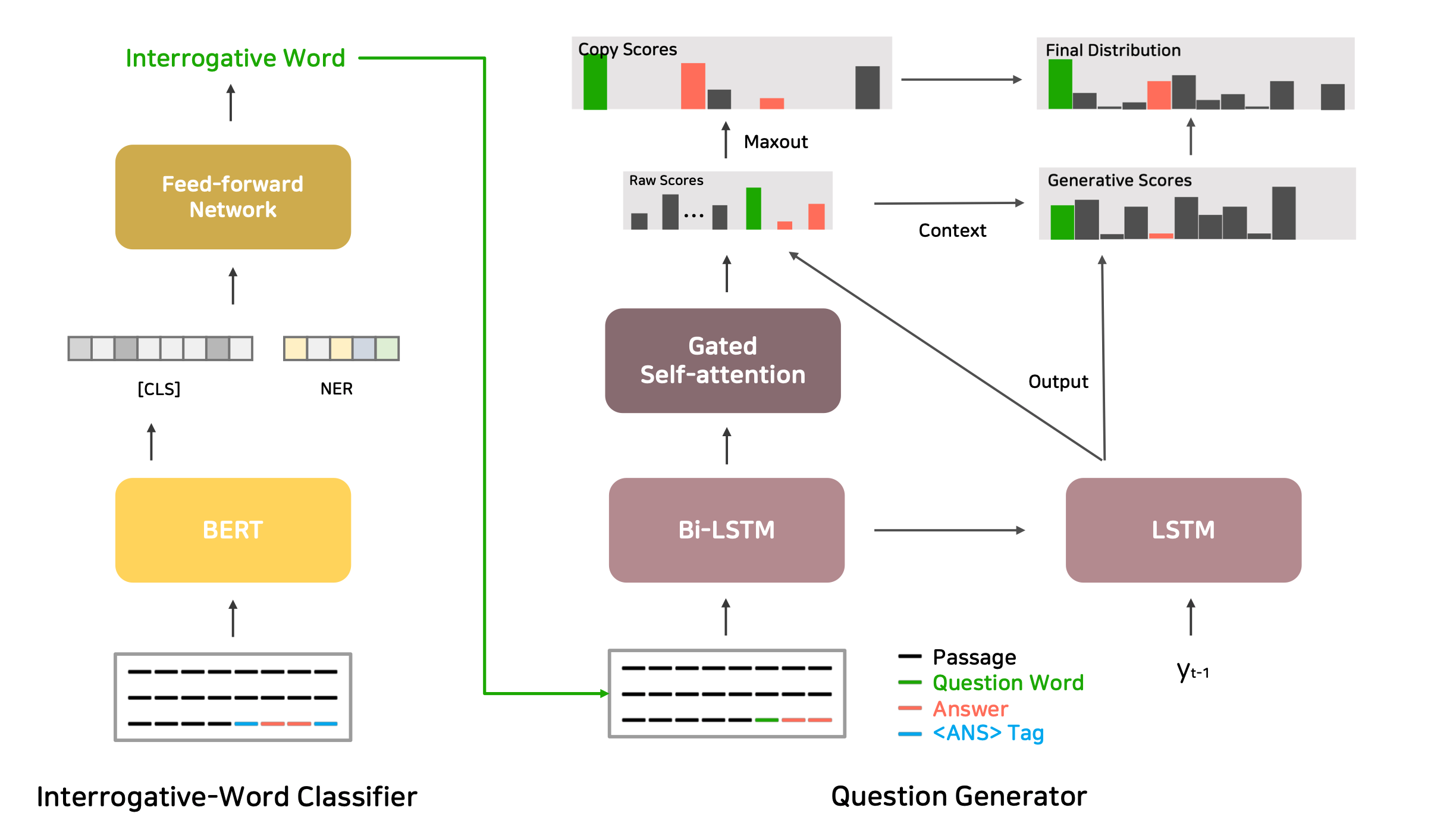}
    \caption{Overall architecture of IWAQG.}
    \label{fig:overall_architecture}
\end{figure*}{}

\section{Interrogative-Word-Aware Question Generation}
\subsection{Problem Statement}
Given a passage $P$, and an answer $A$, we want to find a question $Q$, whose answer is $A$. More formally:

\[
\overline{Q} = \argmax_Q Prob(Q|P,A)
\]

We assume that $P$ is a paragraph composed of a list of words: $P = \{x_t\}_{t=1}^M$, and the answer is a subspan of $P$.

We model this problem with a pipelined approach. First, given $P$ and $A$, we predict the interrogative word $I_w$, and then, we input into QG module $P$, $A$, and $I_w$. The overall architecture of our model is shown in Figure \ref{fig:overall_architecture}.

\subsection{Interrogative-Word Classifier}
As discussed in section \ref{ref:upperbound}, any model can be used to predict interrogative words if its accuracy is high enough. Our interrogative-word classifier is based on BERT, a state-of-the-art model in many NLP tasks that can successfully utilize the context to grasp the semantics of the words inside a sentence \cite{devlin2018bert}. We input a passage that contains the answer of the question we want to build and add the special token \code{[ANS]} to let BERT knows that the answer span has a special meaning and must be used differently to the rest of the passage. As required by BERT, the first token of the input is the special token \code{[CLS]}, and the last is \code{[SEP]}. This \code{[CLS]} token embedding originally was designed for classification tasks. In our case, to classify interrogative words, it learns how to represent the context and the answer information. 

On top of BERT, we build a feed-forward network that receives as input the \code{[CLS]} token embedding concatenated with a learnable embedding of the entity type of the answer, as shown on the left side of Figure \ref{fig:overall_architecture}. We propose to utilize the entity type of the answer because there is a clear correlation between the answer type of the question and the entity type of the answer. For example, if the interrogative word is \textit{who}, the answer is very likely to have an entity type \textit{person}. Since we are using \code{[CLS]} token embedding as a representation of the context and the answer, we consider that using an explicit entity type embedding of the answer could help the system.

\subsection{Question Generator}
For the QG module, we employ one of the current state-of-the-art QG models \cite{zhao2018paragraph}. This model is a sequence-to-sequence neural network that uses a gated self-attention in the encoder and an attention mechanism with maxout pointer in the decoder.

One way to connect the interrogative-word classifier to the QG model is to use the predicted interrogative word as the first output token of the decoder by default. However, we cannot expect a perfect interrogative-word classifier and also, the first word of the questions is not necessarily an interrogative word. Therefore, in this work, we add the predicted interrogative word to the input of the QG model to let the model decide whether to use it or not. In this way, we can condition the generated question on the predicted interrogative word effectively.

\subsubsection{Encoder}
The encoder is composed of a Recurrent Neural Network (RNN), a self-attention network, and a feature fusion gate \cite{gong-bowman-2018-ruminating}. The goal of this fusion gate is to combine two intermediate learnable features into the final encoded passage-answer representation. The input of this model is the passage $P$. It includes the answer and the predicted interrogative word $I_w$, which is located just before the answer span. The RNN receives the word embedding of the tokens of this text concatenated with a learnable meta-embedding that tags if the token is the interrogative word, the answer of the question to generate or the context of the answer.

\subsubsection{Decoder}
The decoder is composed of an RNN with an attention layer and a copy mechanism \cite{Gu_2016}. The RNN of the decoder at time step $t$ receives its hidden state at the previous time step $t-1$ and the previously generated output $y_{t-1}$. At $t = 0$, it receives the last hidden state of the encoder. This model combines the probability of generating a word and the probability of copying that word from the input as shown on the right side of Figure \ref{fig:overall_architecture}. To compute the generative scores, it uses the outputs of the decoder, and the context of the encoder, which is based on the raw attention scores. To compute the copy scores, it uses the outputs of the RNN and the raw attention scores of the encoder. \citet{zhao2018paragraph} observed that the repetition of words in the input sequence tends to create repetitions in the output sequence too. Thus, they proposed a maxout pointer mechanism instead of the regular pointer mechanism \cite{vinyals2015pointer}. This new pointer mechanism limits the magnitude of the scores of the repeated words to their maximum value. To do that, first, the attention scores are computed over the input sequence and then, the score of a word at time step $t$  is calculated as the maximum of all scores pointing to the same word in the input sequence. The final probability distribution is calculated by applying the softmax function on the concatenation of copy scores and generative scores and summing up the probabilities pointing to the same words. 

\section{Experiments}
In our experiments, we study our proposed system on SQuAD dataset v1.1. \cite{Rajpurkar_2016}, prove the validity of our hypothesis and compare it with the current state of the art.

\subsection{Dataset}
In order to train our interrogative-word classifier, we use the training set of SQuAD v1.1 \cite{Rajpurkar_2016}. This dataset is composed of 87599 instances, however, the number of interrogative words is not balanced as seen in \ref{table:squad_stats}. To train the interrogative-word classifier, we downsample the training set to have a balanced dataset.

\begin{table}[H]
    \centering
    \begin{tabular}{|c|c|c|}
        \hline
        Class & Original & After Downsampling \\ \hline
        What & 50385 & 4000 \\ 
        Which & 6111 & 4000 \\ 
        Where & 3731 & 3731\\ 
        When & 5437 & 4000\\ 
        Who & 9162 & 4000 \\ 
        Why & 1224 & 1224\\ 
        How & 9408 & 4000 \\ 
        Others & 9408 & 4000 \\ \hline
        \end{tabular}
    \captionof{table}{SQuAD training set statistics. Full training set and downsampled training set.}
    \label{table:squad_stats}
\end{table}{}

For a fair comparison with previous models, we train the QG model on the training set of SQuAD and split by half the dev set into dev and test randomly as \citet{Zhou2017NeuralQG}.

\subsection{Implementation}
The interrogative-word classifier is made using the PyTorch implementation of BERT-base-uncased made by HuggingFace\footnote{\url{https://github.com/huggingface/pytorch-transformers}}. It was trained for three epochs using cross entropy loss as the objective function. The entity types are obtained using spaCy\footnote{\url{https://spacy.io/}}. If spaCy cannot return an entity for a given answer, we label it as \code{None}. The dimension of the entity type embedding is 5. The input dimension of the classifier is 773 (768 from BERT base hidden size and 5 from the entity type embedding size) and the output dimension is 8 since we predict the interrogative words: \textit{what}, \textit{which}, \textit{where}, \textit{when}, \textit{who}, \textit{why}, \textit{how}, and \textit{others}. The feed-forward network consists of a single layer. For optimization, we used Adam optimizer with weight decay and learning rate of 5e-5. The QG model is based on the model proposed by \cite{zhao2018paragraph} with small modifications using PyTorch. The encoder uses a BiLSTM and the decoder uses an LSTM. During training, the QG model uses the golden interrogative words to enforce the decoder to always copy the interrogative word. On the other hand, during inference, it uses the interrogative word predictions from the classifier.

\begin{table*}[t]
\resizebox{\textwidth}{!}{%
    \begin{tabular}{|c|c|c|c|c|c|c|}
    \hline
    Model & BLEU-1 & BLEU-2 & BLEU-3 & BLEU-4 & METEOR & ROUGE-L \\ \hline
    \citet{Zhou2017NeuralQG} & - & - & - & 13.29 & - & - \\
    \citet{zhao2018paragraph}* & 45.69 & 29.58 & 22.16 & 16.85 & 20.62 & 44.99 \\ 
    \citet{kim2019improving} & - & - & - & 16.17 & - & - \\ 
    \citet{Liu:2019:LGQ:3308558.3313737} & 46.58 & 30.90 & 22.82 & 17.55 & 21.24 & 44.53 \\
    {\textbf{IWAQG}} & {\textbf{47.69}} & {\textbf{32.24}} & {\textbf{24.01}} & {\textbf{18.53}} & {\textbf{22.33}} & {\textbf{46.94}} \\ \hline
    \end{tabular}%
    }
    \caption{Comparison of our model with the baselines. ``*" is our QG module.}
    \label{table:comparison}
\end{table*}

\subsection{Evaluation}
We perform an automatic evaluation using the metrics: BLUE-1, BLUE-2, BLUE-3, BLUE-4 \cite{Papineni:2002:BMA:1073083.1073135}, METEOR \cite{Lavie:2009:MMA:1743627.1743643} and ROUGE-L \cite{Lin-2004-rouge}. In addition, we perform a qualitative analysis where we compare the generated questions of the baseline \cite{zhao2018paragraph}, our proposed model, the upper bound performance of our model, and the golden question.

\section{Results}
\subsection{Comparison with Previous Models}
Our interrogative-word classifier achieves an accuracy of 73.8\% on the test set of SQuAD. Using this model for the pipelined system, we compare the performance of the QG model with respect to the previous state-of-the-art models. Table \ref{table:comparison} shows the evaluation results of our model and the current state-of-the-art models, which are briefly described below.
\begin{itemize}
    \item \citet{Zhou2017NeuralQG} is one of the first authors who proposed a sequence-to-sequence model with attention and copy mechanism. They also proposed the use of POS and NER tags as lexical features for the encoder.
    \item \citet{zhao2018paragraph} proposed the model in which we based our QG module.
    \item \citet{kim2019improving} proposed QG architecture that treats the passage and the target answer separately. 
    \item \citet{Liu:2019:LGQ:3308558.3313737}  proposed a sequence-to-sequence model with a clue word predictor using a Graph Convolutional Networks to identify if each word in the input passage is a potential clue that should be copied into the generated question. 
\end{itemize}{}

Our model outperforms all other models in all the metrics. This improvement is consistent, around 2\%. This is due to the improvement in the recall of the interrogative words. All these measures are based on the overlap between the golden question and the generated question, so using the right interrogative word, we can improve these scores. In addition, generating the right interrogative word also helps to create better questions since the output of the RNN of the decoder at time step $t$ also depends on the previously generated word. 

\subsection{Upper Bound Performance of IWAQG} \label{ref:upperbound}
\begin{figure}[H]
    \centering
    \includegraphics[width=7.5cm]{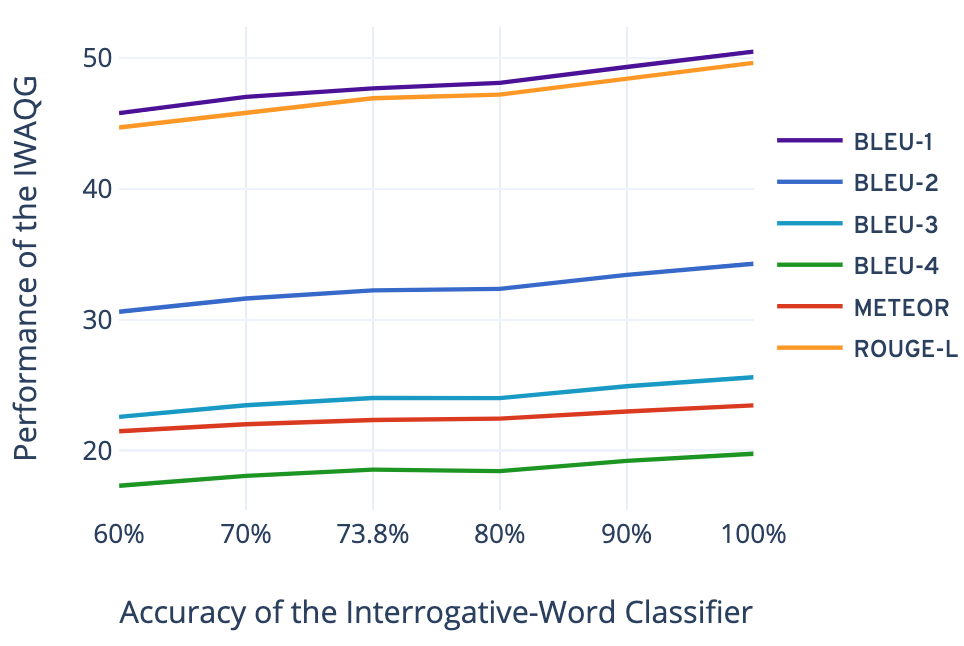}
    \caption{Performance of the QG model with respect to the accuracy of the interrogative-word classifier.}
    \label{fig:upperbound}
\end{figure}

We analyze the upper bound improvement that our QG model can have according to different levels of accuracy of the interrogative-word classifier. In order to do that, instead of using our interrogative-word classifier, we use the golden labels of the test set and generated noise to simulate a classifier with different accuracy levels. Table \ref{table:upperbound} and Figure \ref{fig:upperbound} show a linear relationship between the accuracy of the classifier and the IWAQG. This demonstrates the effectiveness of our pipelined approach regardless of the interrogative-word classifier model. 

\begin{table*}[t]
\resizebox{\textwidth}{!}{%
\begin{tabular}{|c|c|c|c|c|c|c|}
\hline
Accuracy & BLEU-1 & BLEU-2 & BLEU-3 & BLEU-4 & METEOR & ROUGE-L \\ \hline
{\textbf{Only QG*}} & {\textbf{45.63}} & {\textbf{30.43}} & {\textbf{22.51}} & {\textbf{17.30}} & {\textbf{21.06}} & {\textbf{45.42}} \\
60\% & 45.80 & 30.61 & 22.57 & 17.30 & 21.47 & 44.70 \\ 
70\% & 47.05 & 31.62 & 23.46 & 18.05 & 22.00 & 45.88 \\
{\textbf{IWAQG (73.8\%)}} & {\textbf{47.69}} & {\textbf{32.24}} & {\textbf{24.01}} & {\textbf{18.53}} & {\textbf{22.33}} & {\textbf{46.94}} \\
80\% & 48.11 & 32.36 & 24.00 & 18.42 & 22.43 & 47.22 \\
90\% & 49.33 & 33.43 & 24.91 & 19.20 & 22.98 & 48.41 \\
{\textbf{Upper Bound (100\%)}} & {\textbf{50.51}} & {\textbf{34.28}} & {\textbf{25.60}} & {\textbf{19.75}} & {\textbf{23.45}} & {\textbf{49.65}} \\ \hline
\end{tabular}%
}
\caption{Performance of the QG model with respect to the accuracy of the interrogative-word classifier. ``*" is our implementation of the QG module without our interrogative-word classifier \cite{zhao2018paragraph}.}
\label{table:upperbound}
\end{table*}

In addition, we analyze the recall of the interrogative words generated by our pipelined system. As shown in the Table \ref{table:recallQG}, the total recall of using only the QG module is 68.29\%, while the recall of our proposed system, IWAQG, is 74.10\%, an improvement of almost 6\%. Furthermore, if we assume a perfect interrogative-word classifier, the recall would be 99.72\%, a dramatic improvement which proves the validity of our hypothesis.

\begin{table*}[t]
    \centering
    \resizebox{\textwidth}{!}{%
        \begin{tabular}{|c|c|c|c|c|c|c|c|c|c|}
        \hline
        Model & What & Which & Where & When & Who & Why & How & Others & Total \\ \hline
        
        Only QG* & 82.24\% & 0.29\% & 51.90\% & 60.82\% & 68.34\% & 12.66\% & 60.62\% & 2.13\% & 68.29\% \\ 
        
        IWAQG & 87.66\% & 1.46\% & 66.24\% & 49.41\% & 76.41\% & 50.63\% & 70.26\% & 14.89\% & 74.10\% \\ 
        
        Upper Bound & 99.87\% & 99.71\% & 100.00\% & 99.71\% & 99.84\% & 98.73\% & 99.67\% & 89.36\% & 99.72\% \\ \hline
        
        \end{tabular}
    }%

    \caption{Recall of interrogative words of the QG model. ``*" is our implementation of the QG module without our interrogative-word classifier \cite{zhao2018paragraph}.}
    \label{table:recallQG}
\end{table*}{}

\subsection{Effectiveness of the input of interrogative words into the QG model}
\begin{figure}[H]
    \centering
    \includegraphics[width=7.5cm]{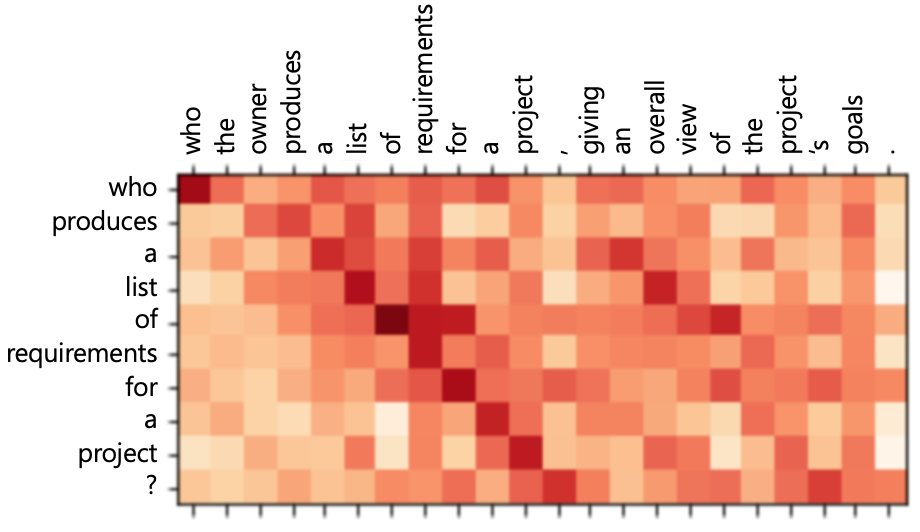}
    \caption{Attention matrix between the generated question (Y-axis) and the given passage (X-axis).}
    \label{fig:attention}
\end{figure}

In this section, we show the effectiveness of inserting explicitly the predicted interrogative word into the passage. We argue that this simple way of connecting the two models exploits the characteristics of the copy mechanism successfully. As we can see in Figure \ref{fig:attention}, the attention score of the generated interrogative word, \textit{who}, is relatively high for the predicted interrogative word and lower for the other words. This means that it is very likely that the interrogative word inserted into the passage is copied as intended.

\subsection{Qualitative Analysis}
In this section, we present a sample of the generated questions of our model, the upper bound model (interrogative-word classifier accuracy is 100\%), the baseline \cite{zhao2018paragraph}, and the golden questions to show how our model improves the recall of the interrogative words with respect to the baseline. In general, our model has a better recall of interrogative words than the baseline which leads us to a better quality of questions. However, since we are still far from a perfect interrogative-word classifier, we also show that questions that our current model cannot generate correctly could be generated well if we had a better classifier.

As we can see in Table \ref{table:qualitative_results}, in the first three examples the interrogative words generated by the baseline are wrong, while our model is right. In addition, due to the wrong selection of interrogative words, in the second example, the topic of the question generated by the baseline is also wrong. On the other hand, since our model selects the right interrogative word, it can create the right question. Each generated word depends on the previously generated word because of the generative LSTM model, so it is very important to select correctly the first word, i.e. the interrogative word. However, the performance of our proposed interrogative-word classifier is not perfect, if it had a 100\% accuracy, then, we could improve the quality of the generated questions like in the last two examples.

\begin{table*}[t]
    \centering
    \resizebox{\textwidth}{!}{%
    \begin{tabular}{m{1em}m{7em}m{7em}m{7em}m{7em}m{5em}}
\toprule

id & Only QG* & IWAQG & Upper Bound & Golden & Answer \\
\midrule
1  & \textcolor{red}{\textbf{what}} produces a list of requirements for a project? & \textcolor{green!70!blue}{\textbf{who}} produces a list of requirements for a project? & \textcolor{green!70!blue}{\textbf{who}} produces a list of requirements for a project? & \textcolor{green!70!blue}{\textbf{who}} produces a list of requirements for a project, giving an overall view of the project's goals?
&  The owner
\\
\midrule
2 & \textcolor{red}{\textbf{how}} many tunnels were constructed through newcastle city centre? & \textcolor{green!70!blue}{\textbf{what}} type of tunnels constructed through newcastle city centre? & \textcolor{green!70!blue}{\textbf{what}} type of tunnels constructed through newcastle city centre ? & \textcolor{green!70!blue}{\textbf{what}} type of tunnels are constructed through newcastle 's city center? & deep-level tunnels
\\
\midrule
3 & \textcolor{red}{\textbf{who}} received a battering during the siege of newcastle? & \textcolor{green!70!blue}{\textbf{what}} received a battering during the siege of newcastle ? & \textcolor{green!70!blue}{\textbf{what}} received a battering during the siege of newcastle ? & \textcolor{green!70!blue}{\textbf{what}} received a battering during the siege of newcastle? & The church tower\\
\midrule
4 &  \textcolor{red}{\textbf{what}} system is newcastle international airport connected to? &  \textcolor{red}{\textbf{what}} system is newcastle international airport connected to? &  \textcolor{green!70!blue}{\textbf{how}} is newcastle international airport connected to ? & \textcolor{green!70!blue}{\textbf{how}} is newport 's airport connected to the city? & via the Metro Light Rail system\\
\midrule
5 & \textcolor{red}{\textbf{who}} was the country most dependent on arab oil? & \textcolor{red}{\textbf{what}} country was the most dependent on arab oil? & \textcolor{green!70!blue}{\textbf{which}} country was the most dependent on arab oil? & \textcolor{green!70!blue}{\textbf{which}} country is the most dependent on arab oil? & Japan \\

\bottomrule
\end{tabular}
    }
    \caption{Qualitative Analysis. Comparison between the baseline, our proposed model, the upper bound of our model, the golden question and the answer of the question. ``*" is our implementation of the QG module without our interrogative-word classifier \cite{zhao2018paragraph}.}
    \label{table:qualitative_results}
\end{table*}{}

\subsection{Ablation Study}
We tried to combine different features shown in Table \ref{table:ablation_study} for the interrogative-word classifier. In this section, we analyze their impact on the performance of the model.

The first model is only using the \code{[CLS]} BERT token embedding \cite{devlin2018bert} that represents the input passage. In this model, the input is the passage where the answer appears but, the model does not know where the answer is. The second model is the previous one with the entity type of the answer as an additional feature. The performance of this model is a bit better than the first one but it is not enough to be utilized effectively for our pipeline. In the third model, the input is the passage. This model uses the average of the answer token embeddings generated by BERT along with the \code{[CLS]} token embedding. As we can see, the performance noticeably increased, which indicates that answer information is the key to predict the interrogative word needed. In the fourth model, we added the special token \code{[ANS]} at the beginning and at the end of the answer span to let BERT knows where the answer is in the passage. So the input to the feed-forward network is only the \code{[CLS]} token embedding. This model clearly outperforms the previous one, which shows that BERT can exploit the answer information better if it is tagged with the \code{[ANS]} token. The fifth model is the same as the previous one but with the addition of the entity-type embedding of the answer. The combination of the three features (answer, answer entity type, and passage) yields to the best performance.

\begin{table}[H]
    \centering
        \begin{tabular}{|c|c|}
            \hline
            Classifier & Accuracy \\ 
            \hline
            CLS & 56.0\% \\ 
            CLS + NER & 56.6\% \\ 
            CLS + AE & 70.3\% \\ 
            CLS + AT & 73.3\% \\ 
            \textbf{CLS + AT + NER} & \textbf{73.8\%} \\
            \hline
    \end{tabular}
    \caption{Ablation Study of our interrogative-word classifier.}
    \label{table:ablation_study}
    \vspace{-4mm}
\end{table}{}

In addition, we provide the recall and precision per class for our final interrogative-word classifier (CLS + AT + NER in Table \ref{table:model1precision}). As we can see, the overall recall is high, and it is also higher than just using the QG module (Table \ref{table:recallQG}), which proves our hypothesis that modeling the interrogative-word prediction task as an independent classification problem yields to a higher recall than generating them with the full question. However, the recall of \textit{which} is very low. This is due to the intrinsic difficulty of predicting this interrogative words. Questions like ``what country" and ``which country" can be correct depending on the context, but the meaning is very similar. Our model has also problem with \textit{why} due to the lack of training instances for this class. Lastly, the recall of `\textit{when} is also low because many questions of this type can be formulated with other interrogative words, e.g.: instead of ``When did WWII start?", we can ask ``In which year did WWII start?".

\begin{table}[H]
    \centering
    \begin{tabular}{|c|c|c|}
        \hline
        Class & Recall & Precision \\ \hline
        What & 87.7\% & 76.0\% \\ 
        Which & 1.4\% & 38.0\% \\ 
        Where & 65.9\% & 55.8\% \\ 
        When & 49.2\% & 69.8\% \\  
        Who & 76.9\% & 66.7\% \\ 
        Why & 50.1\% & 74.1\% \\ 
        How & 70.5\% & 79.0\% \\ 
        Others & 10.5\% & 57.0\% \\ \hline
        \end{tabular}
    \captionof{table}{Recall and precision of interrogative words of our interrogative-word classifier.}
    \label{table:model1precision}
\end{table}{}

\section{Conclusion and Future Work}
In this work, we proposed an Interrogative-Word-Aware Question Generation (IWAQG), a pipelined model composed of an interrogative-word classifier and a question generator to tackle the question generation task. First, we predict the interrogative word. Then, the Question Generation (QG) model generates the question using the predicted interrogative word. Thanks to this independent interrogative-word classifier and the copy mechanism of the question generation model, we are able to improve the recall of the interrogative words in the generated questions. This improvement also leads to a better quality of the generated questions. We prove our hypotheses through quantitative and qualitative experiments, showing that our pipelined system outperforms the previous state-of-the-art models. Lastly, we also prove that our methodology is remarkably effective, showing a theoretical upper bound of the potential improvement using a more accurate interrogative-word classifier.

In the future, we would like to improve the interrogative-word classifier, since it would clearly improve the performance of the whole system as we showed. We also expect that the use of the Transformer architecture\cite{vaswani2017attention} could improve the QG model. In addition, we plan to test our approach on other datasets to prove its generalization capability. Finally, an interesting application of this work could be to utilize QG to improve Question Answering systems.

\section*{Acknowledgements}
This research was supported by Next-Generation Information Computing Development Program through the National Research Foundation of Korea (NRF) funded by the Ministry of Science and ICT (2017M3C4A7065962).

\bibliography{emnlp-ijcnlp-2019}

\begin{thebibliography}{23}
\expandafter\ifx\csname natexlab\endcsname\relax\def\natexlab#1{#1}\fi

\bibitem[{Devlin et~al.(2018)Devlin, Chang, Lee, and
  Toutanova}]{devlin2018bert}
Jacob Devlin, Ming-Wei Chang, Kenton Lee, and Kristina Toutanova. 2018.
\newblock Bert: Pre-training of deep bidirectional transformers for language
  understanding.
\newblock \emph{arXiv preprint arXiv:1810.04805}.

\bibitem[{Du et~al.(2017)Du, Shao, and Cardie}]{du2017learning}
Xinya Du, Junru Shao, and Claire Cardie. 2017.
\newblock Learning to ask: Neural question generation for reading
  comprehension.
\newblock In \emph{Association for Computational Linguistics (ACL)}.

\bibitem[{Duan et~al.(2017)Duan, Tang, Chen, and Zhou}]{Duan2017QuestionGF}
Nan Duan, Duyu Tang, Peng Chen, and Ming Zhou. 2017.
\newblock Question generation for question answering.
\newblock In \emph{EMNLP}.

\bibitem[{Gong and Bowman(2018)}]{gong-bowman-2018-ruminating}
Yichen Gong and Samuel Bowman. 2018.
\newblock \href {https://doi.org/10.18653/v1/W18-2601} {Ruminating reader:
  Reasoning with gated multi-hop attention}.
\newblock In \emph{Proceedings of the Workshop on Machine Reading for Question
  Answering}, pages 1--11, Melbourne, Australia. Association for Computational
  Linguistics.

\bibitem[{Gu et~al.(2016)Gu, Lu, Li, and Li}]{Gu_2016}
Jiatao Gu, Zhengdong Lu, Hang Li, and Victor~O.K. Li. 2016.
\newblock \href {https://doi.org/10.18653/v1/p16-1154} {Incorporating copying
  mechanism in sequence-to-sequence learning}.
\newblock \emph{Proceedings of the 54th Annual Meeting of the Association for
  Computational Linguistics (Volume 1: Long Papers)}.

\bibitem[{Heilman and Smith(2010)}]{heilman-smith-2010-good}
Michael Heilman and Noah~A. Smith. 2010.
\newblock \href {https://www.aclweb.org/anthology/N10-1086} {Good question!
  statistical ranking for question generation}.
\newblock In \emph{Human Language Technologies: The 2010 Annual Conference of
  the North {A}merican Chapter of the Association for Computational
  Linguistics}, pages 609--617, Los Angeles, California. Association for
  Computational Linguistics.

\bibitem[{Joshi et~al.(2017)Joshi, Choi, Weld, and Zettlemoyer}]{Joshi_2017}
Mandar Joshi, Eunsol Choi, Daniel Weld, and Luke Zettlemoyer. 2017.
\newblock \href {https://doi.org/10.18653/v1/p17-1147} {Triviaqa: A large scale
  distantly supervised challenge dataset for reading comprehension}.
\newblock \emph{Proceedings of the 55th Annual Meeting of the Association for
  Computational Linguistics (Volume 1: Long Papers)}.

\bibitem[{Kim et~al.(2019)Kim, Lee, Shin, and Jung}]{kim2019improving}
Yanghoon Kim, Hwanhee Lee, Joongbo Shin, and Kyomin Jung. 2019.
\newblock Improving neural question generation using answer separation.
\newblock In \emph{Proceedings of the AAAI Conference on Artificial
  Intelligence}, volume~33, pages 6602--6609.

\bibitem[{Labutov et~al.(2015)Labutov, Basu, and
  Vanderwende}]{Labutov2015DeepQW}
Igor Labutov, Sumit Basu, and Lucy Vanderwende. 2015.
\newblock \href {https://doi.org/10.3115/v1/P15-1086} {Deep questions without
  deep understanding}.
\newblock In \emph{Proceedings of the 53rd Annual Meeting of the Association
  for Computational Linguistics and the 7th International Joint Conference on
  Natural Language Processing}, pages 889--898.

\bibitem[{Lavie and Denkowski(2009)}]{Lavie:2009:MMA:1743627.1743643}
Alon Lavie and Michael~J. Denkowski. 2009.
\newblock \href {https://doi.org/10.1007/s10590-009-9059-4} {The meteor metric
  for automatic evaluation of machine translation}.
\newblock \emph{Machine Translation}, 23(2-3):105--115.

\bibitem[{Lewis et~al.(2019)Lewis, Denoyer, and
  Riedel}]{lewis2019unsupervisedqa}
Patrick Lewis, Ludovic Denoyer, and Sebastian Riedel. 2019.
\newblock Unsupervised question answering by cloze translation.
\newblock In \emph{Proceedings of the 57th Annual Meeting of the Association
  for Computational Linguistics (Volume 1: Long Papers)}.

\bibitem[{Lin(2004)}]{Lin-2004-rouge}
Chin-Yew Lin. 2004.
\newblock \href {https://www.aclweb.org/anthology/W04-1013} {{ROUGE}: A package
  for automatic evaluation of summaries}.
\newblock In \emph{Text Summarization Branches Out}, pages 74--81, Barcelona,
  Spain. Association for Computational Linguistics.

\bibitem[{Liu et~al.(2019)Liu, Zhao, Niu, Lai, He, Wei, and
  Xu}]{Liu:2019:LGQ:3308558.3313737}
Bang Liu, Mingjun Zhao, Di~Niu, Kunfeng Lai, Yancheng He, Haojie Wei, and
  Yu~Xu. 2019.
\newblock \href {https://doi.org/10.1145/3308558.3313737} {Learning to generate
  questions by learningwhat not to generate}.
\newblock In \emph{The World Wide Web Conference}, WWW '19, pages 1106--1118,
  New York, NY, USA. ACM.

\bibitem[{Mazidi and Nielsen(2014)}]{qgrules}
Karen Mazidi and Rodney Nielsen. 2014.
\newblock \href {https://doi.org/10.3115/v1/P14-2053} {Linguistic
  considerations in automatic question generation}.
\newblock In \emph{52nd Annual Meeting of the Association for Computational
  Linguistics, ACL 2014 - Proceedings of the Conference}, volume~2.

\bibitem[{Pan et~al.(2019)Pan, Lei, Chua, and Kan}]{pan2019recent}
Liangming Pan, Wenqiang Lei, Tat-Seng Chua, and Min-Yen Kan. 2019.
\newblock Recent advances in neural question generation.
\newblock \emph{arXiv preprint arXiv:1905.08949}.

\bibitem[{Papineni et~al.(2002)Papineni, Roukos, Ward, and
  Zhu}]{Papineni:2002:BMA:1073083.1073135}
Kishore Papineni, Salim Roukos, Todd Ward, and Wei-Jing Zhu. 2002.
\newblock \href {https://doi.org/10.3115/1073083.1073135} {Bleu: A method for
  automatic evaluation of machine translation}.
\newblock In \emph{Proceedings of the 40th Annual Meeting on Association for
  Computational Linguistics}, ACL '02, pages 311--318, Stroudsburg, PA, USA.
  Association for Computational Linguistics.

\bibitem[{Rajpurkar et~al.(2016)Rajpurkar, Zhang, Lopyrev, and
  Liang}]{Rajpurkar_2016}
Pranav Rajpurkar, Jian Zhang, Konstantin Lopyrev, and Percy Liang. 2016.
\newblock \href {https://doi.org/10.18653/v1/d16-1264} {Squad: 100,000+
  questions for machine comprehension of text}.
\newblock \emph{Proceedings of the 2016 Conference on Empirical Methods in
  Natural Language Processing}.

\bibitem[{Subramanian et~al.(2018)Subramanian, Wang, Yuan, Zhang, Trischler,
  and Bengio}]{subramanian2018neural}
Sandeep Subramanian, Tong Wang, Xingdi Yuan, Saizheng Zhang, Adam Trischler,
  and Yoshua Bengio. 2018.
\newblock Neural models for key phrase extraction and question generation.
\newblock In \emph{Proceedings of the Workshop on Machine Reading for Question
  Answering}, pages 78--88.

\bibitem[{Sun et~al.(2018)Sun, Liu, Lyu, He, Ma, and
  Wang}]{sun-etal-2018-answer}
Xingwu Sun, Jing Liu, Yajuan Lyu, Wei He, Yanjun Ma, and Shi Wang. 2018.
\newblock \href {https://doi.org/10.18653/v1/D18-1427} {Answer-focused and
  position-aware neural question generation}.
\newblock In \emph{Proceedings of the 2018 Conference on Empirical Methods in
  Natural Language Processing}, pages 3930--3939, Brussels, Belgium.
  Association for Computational Linguistics.

\bibitem[{Vaswani et~al.(2017)Vaswani, Shazeer, Parmar, Uszkoreit, Jones,
  Gomez, Kaiser, and Polosukhin}]{vaswani2017attention}
Ashish Vaswani, Noam Shazeer, Niki Parmar, Jakob Uszkoreit, Llion Jones,
  Aidan~N Gomez, {\L}ukasz Kaiser, and Illia Polosukhin. 2017.
\newblock Attention is all you need.
\newblock In \emph{Advances in neural information processing systems}, pages
  5998--6008.

\bibitem[{Vinyals et~al.(2015)Vinyals, Fortunato, and
  Jaitly}]{vinyals2015pointer}
Oriol Vinyals, Meire Fortunato, and Navdeep Jaitly. 2015.
\newblock Pointer networks.
\newblock In \emph{Advances in Neural Information Processing Systems}, pages
  2692--2700.

\bibitem[{Zhao et~al.(2018)Zhao, Ni, Ding, and Ke}]{zhao2018paragraph}
Yao Zhao, Xiaochuan Ni, Yuanyuan Ding, and Qifa Ke. 2018.
\newblock Paragraph-level neural question generation with maxout pointer and
  gated self-attention networks.
\newblock In \emph{Proceedings of the 2018 Conference on Empirical Methods in
  Natural Language Processing}, pages 3901--3910.

\bibitem[{Zhou et~al.(2017)Zhou, Yang, Wei, Tan, Bao, and
  Zhou}]{Zhou2017NeuralQG}
Qingyu Zhou, Nan Yang, Furu Wei, Chuanqi Tan, Hangbo Bao, and Ming Zhou. 2017.
\newblock Neural question generation from text: A preliminary study.
\newblock In \emph{NLPCC}.

\end{thebibliography}
\bibliographystyle{acl_natbib}

\end{document}